\documentclass[conference]{IEEEtran}
\IEEEoverridecommandlockouts

\usepackage{cite}
\usepackage{multicol}
\usepackage{multirow}
\usepackage{amsmath,amssymb,amsfonts}
\usepackage{subcaption}
\usepackage{algorithmic}
\usepackage{graphicx}
\usepackage{textcomp}
\usepackage{xcolor}
\usepackage{todonotes}
\usepackage[hidelinks]{hyperref}
\usepackage{balance}
\usepackage{tcolorbox}

\def\BibTeX{{\rm B\kern-.05em{\sc i\kern-.025em b}\kern-.08em
    T\kern-.1667em\lower.7ex\hbox{E}\kern-.125emX}}
\begin{document}

\begin{titlepage}
\begin{center}
\vspace*{10pt}
\textbf{\Large D-HUMOR: \textbf{D}ark \textbf{H}umor \textbf{U}nderstanding via \textbf{M}ultimodal \textbf{O}pen-ended \textbf{R}easoning \\- A Benchmark Dataset and Method}
\vspace*{20pt}

\begin{flushleft}
Sai Kartheek Reddy Kasu$^\dagger$ (Indian Institute of Information Technology Dharwad, India)\\
Mohammad Zia Ur Rehman$^\dagger$ (Indian Institute of Technology Indore, India)\\ 
Shahid Shafi Dar (Indian Institute of Technology Indore, India)\\ Rishi Bharat Junghare (Indian Institute of Technology Indore, India)\\
Dhanvin Sanjay Namboodiri (Malaviya National Institute of Technology Jaipur, India)\\Nagendra Kumar$^*$ (Indian Institute of Technology Indore, India) \\

\hspace{1pt}

\textbf{Email(s):} 21bds027@iiitdwd.ac.in, phd2101201005@iiti.ac.in, phd2201201004@iiti.ac.in, cse220001063@iiti.ac.in, 2022uec1253@mnit.ac.in, nagendra@iiti.ac.in\\

\vspace{10mm}
$^\dagger$These are co-first authors with equal contribution.\\
$^*$Corresponsding author\\

\vspace{2cm}
\normalsize
This is the preprint version of the accepted paper.\\
\textbf{Accepted in \textit{IEEE International Conference on Data Mining (ICDM)}, 2025} \\

\end{flushleft}        
\end{center}
\end{titlepage}

\title{
D-HUMOR: \textbf{D}ark \textbf{H}umor \textbf{U}nderstanding via \textbf{M}ultimodal \textbf{O}pen-ended \textbf{R}easoning - A Benchmark Dataset and Method\\
}



\author{
\IEEEauthorblockN{Sai Kartheek Reddy Kasu$^\dagger$}
\IEEEauthorblockA{\textit{Indian Institute of Information Technology Dharwad, India} \\
21bds027@iiitdwd.ac.in}

\vspace{1.2em} 

\IEEEauthorblockN{Shahid Shafi Dar}
\IEEEauthorblockA{\textit{Indian Institute of Technology Indore, India} \\
phd2201201004@iiti.ac.in}

\vspace{1.2em} 
\IEEEauthorblockN{Dhanvin Sanjay Namboodiri}
\IEEEauthorblockA{\textit{Malaviya National Institute of Technology Jaipur, India} \\
2022uec1253@mnit.ac.in}

\and

\IEEEauthorblockN{Mohammad Zia Ur Rehman$^\dagger$}
\IEEEauthorblockA{\textit{Indian Institute of Technology Indore, India} \\
phd2101201005@iiti.ac.in}

\vspace{1.2em}

\IEEEauthorblockN{Rishi Bharat Junghare}
\IEEEauthorblockA{\textit{Indian Institute of Technology Indore, India} \\
cse220001063@iiti.ac.in}

\vspace{1.2em}

\IEEEauthorblockN{Nagendra Kumar$^*$}
\IEEEauthorblockA{\textit{Indian Institute of Technology Indore, India} \\
nagendra@iiti.ac.in}

\thanks{$^\dagger$These are co-first authors with equal contribution.}
\thanks{$^*$Corresponding author}
}

\maketitle
\begin{abstract}
Dark humor in online memes poses unique challenges due to its reliance on implicit, sensitive, and culturally contextual cues. To address the lack of resources and methods for detecting dark humor in multimodal content, we introduce a novel dataset of 4,379 Reddit memes annotated for dark humor, target category (gender, mental health, violence, race, disability, and other), and a three-level intensity rating (mild, moderate, severe). Building on this resource, we propose a reasoning-augmented framework that first generates structured explanations for each meme using a Large Vision–Language Model (VLM). Through a Role-Reversal Self-Loop, VLM adopts the author’s perspective to iteratively refine its explanations, ensuring completeness and alignment. We then extract textual features from both the OCR transcript and the self-refined reasoning via a text encoder, while visual features are obtained using a vision transformer. A Tri‐stream Cross‐Reasoning Network (TCRNet) fuses these three streams, text, image, and reasoning, via pairwise attention mechanisms, producing a unified representation for classification. Experimental results demonstrate that our approach outperforms strong baselines across three tasks: dark humor detection, target identification, and intensity prediction. The dataset, annotations, and code are released to facilitate further research in multimodal humor understanding and content moderation.\\
\noindent \textbf{Code and Dataset Access:} \href{https://github.com/Sai-Kartheek-Reddy/D-Humor-Dark-Humor-Understanding-via-Multimodal-Open-ended-Reasoning}{https://github.com/Sai-Kartheek-Reddy/D-Humor-Dark-Humor-Understanding-via-Multimodal-Open-ended-Reasoning}

\end{abstract}


\noindent \textcolor{red}{\textbf{Caution:} This paper contains offensive content due to the nature of the topic, which may be disturbing or offensive to some readers. Reader discretion is advised.}
\vspace{-10pt}
\section{Introduction}
\begin{figure}[!ht]
\centering
\begin{subfigure}{0.48\columnwidth}
\includegraphics[width=\linewidth, height=3cm]{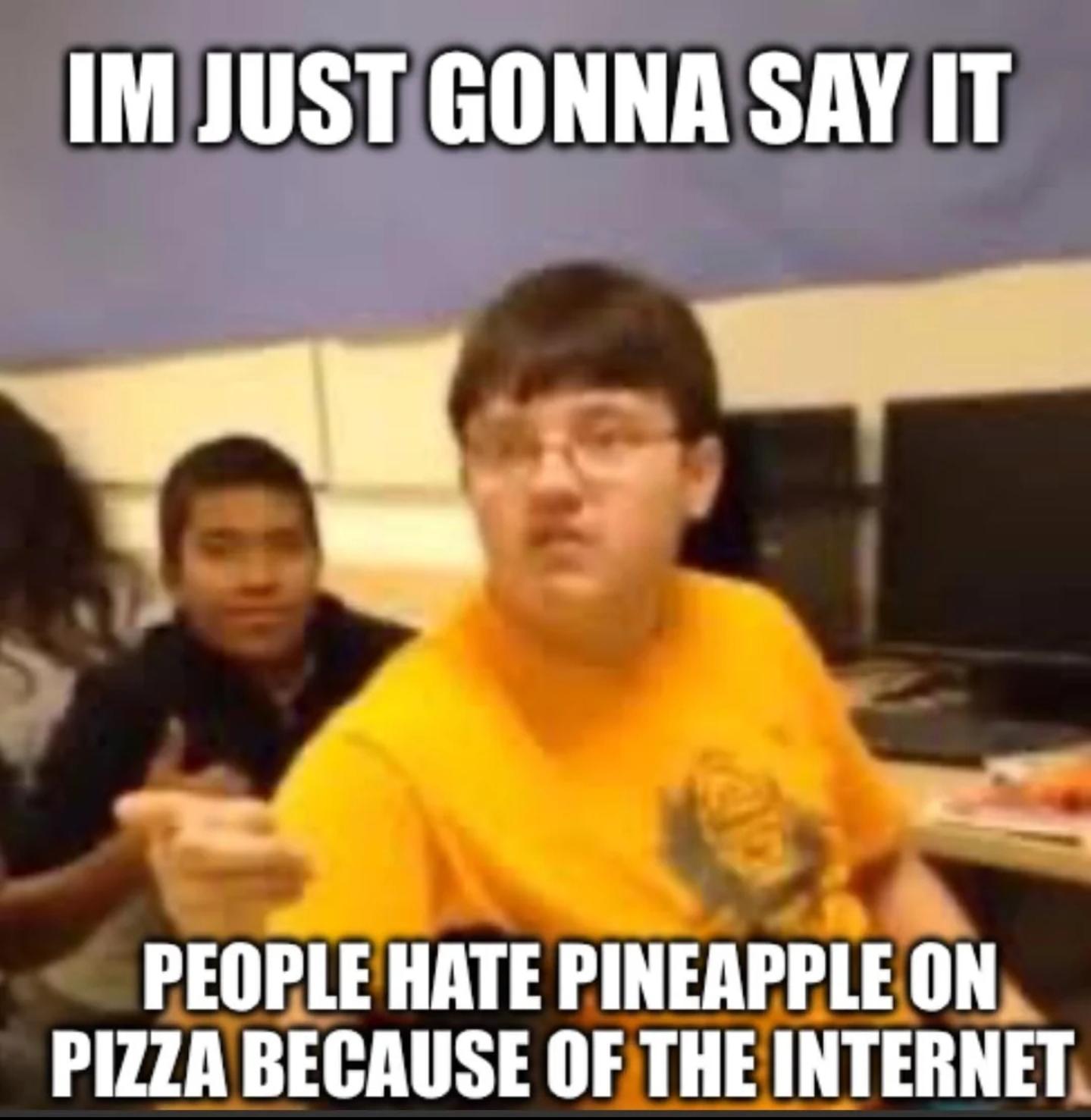} 
\caption{Clean humor}
\label{fig:clean1}
\end{subfigure}
\hfill
\begin{subfigure}{0.48\columnwidth}
\includegraphics[width=\linewidth, height=3cm]{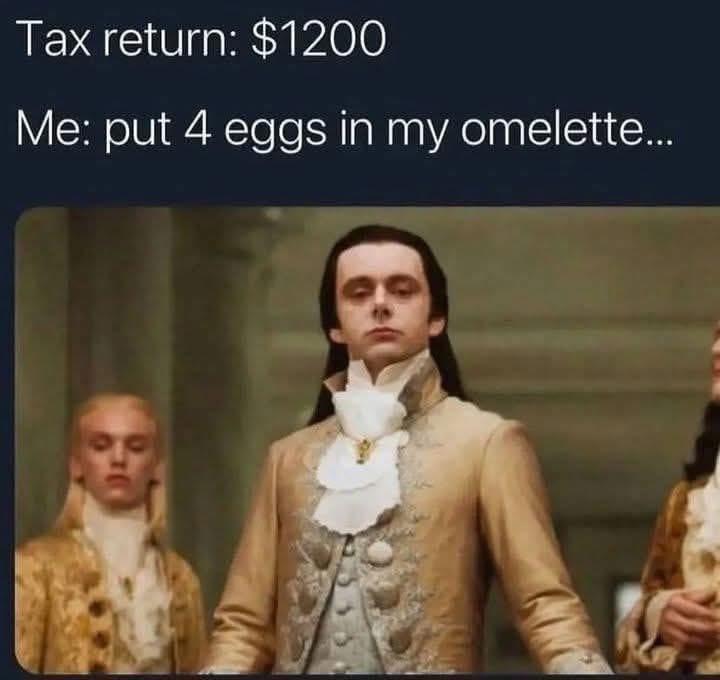} 
\caption{Clean humor}
\label{fig:clean2}
\end{subfigure}

\vspace{0.5em} 

\begin{subfigure}{0.48\columnwidth}
\includegraphics[width=\linewidth, height=3cm]{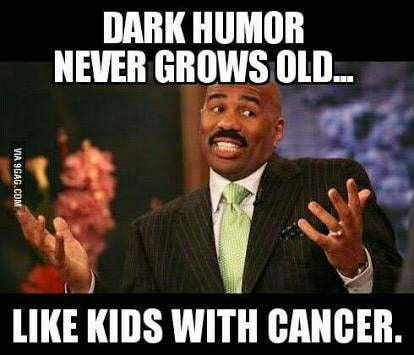}
\caption{Dark humor}
\label{fig:dark1}
\end{subfigure}
\hfill
\begin{subfigure}{0.48\columnwidth}
\includegraphics[width=\linewidth, height=3cm]{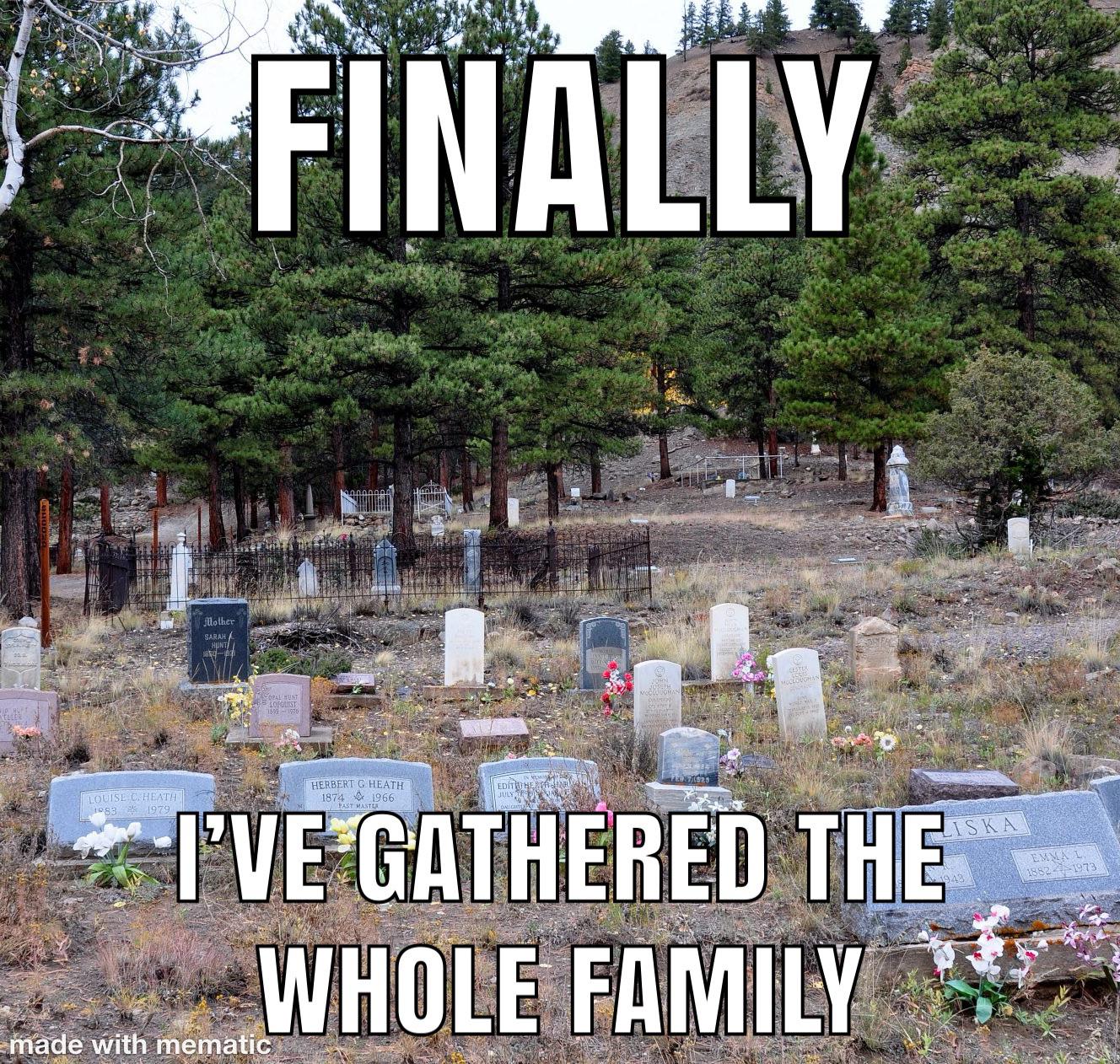}
\caption{Dark humor}
\label{fig:dark2}
\end{subfigure}

\caption{A comparison of clean and dark humor (Source: Reddit).}
\label{fig:memes}
\end{figure}

Dark humor represents a challenging domain for data mining, as it uses subtle elements to convey complex and often controversial forms of humor. The difference between conventional ``clean" humor and dark humor can be quite subtle. While there can never be an objective way to distinguish them, there are some key identifiers for dark humor. Clean humor generally uses topics that everyone can relate to and collectively laugh at. On the other hand, dark humor exploits topics that are typically considered insensitive, as shown in \autoref{fig:memes}.
According to Ruch et al. \cite{ruch2018broadening}, dark humor typically incorporates one of four distinct elements, sarcasm, cynicism, satire, or irony, each characterized by unique linguistic and cognitive features that shape its expression.

However, in today’s digital landscape, social media platforms frequently convey dark humor not just through linguistic expressions but also by image-text combinations \cite{li2024attention,xi2025multimodal} such as memes, which add a multimodal layer to their interpretation. While memes have become a dominant form of online communication, detecting dark humor within them remains an underexplored problem due to the interplay between imagery, text, and cultural context. Effectively mining and detecting dark humor memes requires multimodal approaches capable of integrating textual and visual cues to accurately capture these subtle humor styles \cite{rahman2025camfusion,singh2025ramm}. Advancing automated detection in this space can significantly enhance content moderation, sentiment analysis, and the broader understanding of how humor operates in complex, multimodal social media data~\cite{rehman2025multimodal}. However, to the best of our knowledge, there exists no multimodal dark humor dataset, which limits progress in this underexplored yet critical area.

To address the scarcity of resources for studying dark humor in memes, we introduce a novel dataset comprising a large collection of memes sourced from popular social media platforms, each containing both visual and textual components for comprehensive multimodal analysis. Annotators annotated the dataset not only to distinguish between dark humor and non-dark humor memes in a binary classification framework but also to identify target groups commonly referenced or affected by dark humor, including gender, mental health, violence, race, disability, and other sensitive categories. This granular labeling supports a deeper exploration of the social and ethical dimensions embedded within dark humor, enabling both the technical challenge of humor detection and the analysis of its underlying themes and potential impacts in online meme culture. Since humor plays an important role in mental health \cite{szabo2003acute,basler2025aren}, our dataset could be an important resource.

In addition to the dataset, we propose a novel multimodal approach for dark humor detection in memes that harnesses the complementary strengths of large language models (LLMs)\cite{brown2020language} and visual‐textual feature extraction. Specifically, we first employ the Qwen-2.5-7B Vision–Language Model to generate an initial structured reasoning for each meme, capturing high‐level contextual and inferential cues. Next, we apply an Iterative Reasoning Refinement via Role-Reversal Self-Loop. Qwen is prompted to “become the original author” of the meme and critically review its own explanation, verifying that each field (summary, implied joke, narrative structure, emotional effect, dark attributes, and target) is both faithful and complete. This role-reversal step mimics reflective human reasoning, encouraging the model to correct omissions or misalignments before yielding a final, refined explanation. Textual features are then extracted from both the meme’s transcript using BERT and the self-refined reasoning using S-BERT, while visual features are obtained via ViT for multimodal representation learning. To effectively model the interplay among text, image, and reasoning, we employ a Tri‐stream Cross‐Reasoning Network (TCRNet) that applies pairwise scaled dot‐product attention across all three modalities. The resulting attended representations are concatenated into a unified feature vector, which is passed to a final classification head. This pipeline, combining self-refined reasoning with multimodal fusion, demonstrates the power of hybrid, alignment-aware architectures in capturing the nuanced affective and figurative characteristics of dark humor in social media memes.

Key Contributions of this work are as follows:
\begin{itemize}
    \item \textbf{Dark Humor Meme Dataset:} We introduce a first multimodal dark humor dataset consisting of memes collected from Reddit. Each entry includes both image and OCR-extracted text, annotated not only for binary \emph{dark humor} versus \emph{non-dark humor}, but also for \emph{target category} ( gender, mental health, violence, race, disability, and other) and a three-level \emph{intensity} rating (mild, moderate, severe). This granular annotation enables detailed analysis of social and ethical dimensions in dark humor in memes.

    \item \textbf{Iterative Self-Refined Reasoning:} We use Qwen-2.5-7B VLM to generate an initial structured explanation (summary, implied joke, narrative, emotional effect, dark attributes, target). Then, via a \emph{Role-Reversal Self-Loop}, Qwen assumes the meme author’s perspective to critique and refine its own reasoning, ensuring higher alignment and completeness before feature extraction.

\item \textbf{Reasoning-Enhanced Multimodal Framework:} We propose a multimodal architecture that fuses image embeddings, text embeddings, and embeddings of the self-refined Qwen reasoning through a Tri-stream Cross-Reasoning Network (TCRNet). Pairwise attention across the three streams captures implicit and semantic cues, yielding state-of-the-art performance on dark humor detection, target identification, and intensity prediction.

\end{itemize}

\section{Literature Review}
Dark humor, while increasingly present in online discourse, remains underexplored in computational linguistics. Unlike general humor detection, dark humor poses unique challenges due to its reliance on morally deviant, taboo, or emotionally provocative content. Ruch et al. \cite{ruch2018broadening} identify four rhetorical styles central to dark humor, sarcasm, cynicism, satire, and irony, each involving complex linguistic and psychological cues. While studies directly targeting dark humor are rare, significant advances have been made in detecting its constituent forms.

Sarcasm detection has seen considerable progress, particularly through multimodal approaches. Bao et al. \cite{bao2025hian} proposed HIAN, which uses BiLSTM, attention, and transformers to fuse text, image, and class-based features, capturing cross-modal inconsistencies indicative of sarcasm. Gupta et al. \cite{gupta2025novel} extended this with MLFTI, a weighted multimodal fusion model combining text and visual cues in a shared latent space, achieving 93\% accuracy on Twitter data. Jia et al. \cite{jia2024debiasing} addressed generalization challenges with DMSD-CL, which uses counterfactual data augmentation and contrastive learning to reduce bias from spurious correlations. Liang et al. \cite{liang2024fusion} introduced MMGCL, a graph-based contrastive learning model that integrates text, OCR, and visual objects, effectively modeling sentiment incongruity, an important signal in sarcastic and darkly humorous content.

Irony detection has also been actively explored. Saroj and Pal \cite{saroj2024ensemble} demonstrated the importance of domain adaptation with an ensemble of contextual models fine-tuned for political discourse. Maladry et al. \cite{maladry2024limitations} achieved strong results in Dutch by combining transformer models with engineered sentiment-clash features. In a low-resource setting, Bölücü and Can \cite{bolucu2025semantically} employed UCCA-based graph neural networks for Turkish irony detection, achieving an F1-score of 94.85\%, and demonstrating the value of structured semantic modeling.

Satire, though less studied, has seen recent computational advances. Goyal et al. \cite{goyal2024multimodal} contrasted satire with sarcasm in a multimodal framework, combining BERT, ResNet, and I3D to capture linguistic and visual nuances. Ortega-Bueno et al. \cite{ortega2022multi} proposed MvAttLSTM, a multilingual model that integrates linguistic, sentence-level, and contextual embeddings to detect satire and irony in Spanish. Kamal et al. \cite{kamal2024contextualized} introduced BiSAT, a BiLSTM with self-attention and handcrafted linguistic features, achieving strong results on short satirical texts across multiple datasets.

While these studies provide a robust foundation, several limitations remain. Most notably, research tends to treat sarcasm, irony, satire, and related styles as isolated tasks, overlooking the fact that they frequently co-occur or interact within a single instance of humor, particularly in dark humor. Though there are a few works \cite{ogbanufe2025job, liu2025seeing}, cynicism remains virtually unexamined as a computational task. Furthermore, many models are tailored to narrow domains or specific modalities, limiting their generalizability across contexts, cultures, and moral boundaries. These gaps highlight the necessity of framing sarcasm, irony, satire, and cynicism not as independent phenomena but as interrelated components of a broader category, dark humor. A unified modeling approach is essential to capture the layered, morally provocative, and affectively ambiguous nature of dark humor in real-world discourse.

\section{Dataset Description and Annotation Process}

In this section, we describe the process of dataset collection and the annotation methodology employed. We detail the sources from which the data is gathered, the criteria used for selecting samples, and the steps taken to ensure annotation quality and consistency. This comprehensive overview provides the foundation for understanding the dataset’s structure and reliability for subsequent analysis.

\begin{figure*}[t]
\centering

\begin{subfigure}{0.24\textwidth}
\includegraphics[width=\linewidth, height=3.5cm]{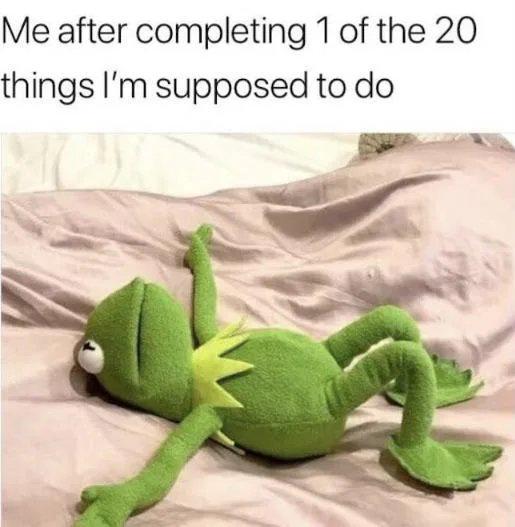}
\caption{Dark Humor: No\\Target: N/A\\Intensity: N/A}
\label{fig:img1}
\end{subfigure}
\hfill
\begin{subfigure}{0.24\textwidth}
\includegraphics[width=\linewidth, height=3.5cm]{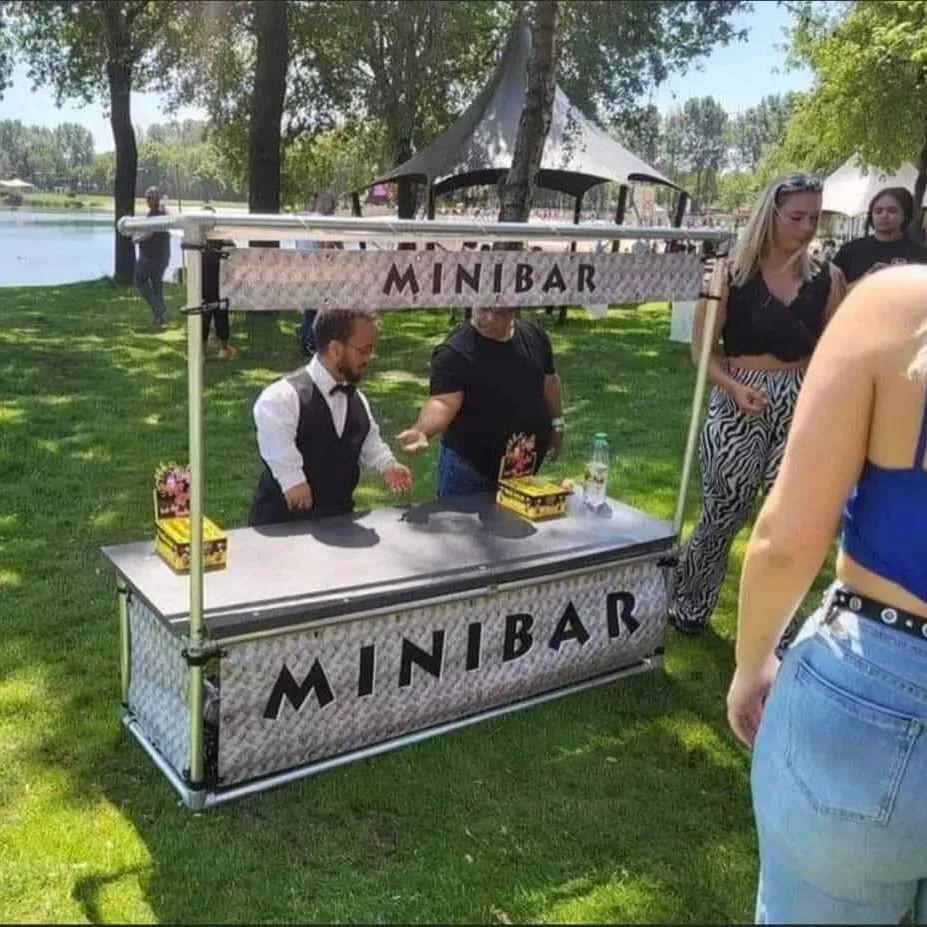}
\caption{Dark Humor: Yes\\Target: Disability\\Intensity: 2}
\label{fig:img2}
\end{subfigure}
\hfill
\begin{subfigure}{0.24\textwidth}
\includegraphics[width=\linewidth, height=3.5cm]{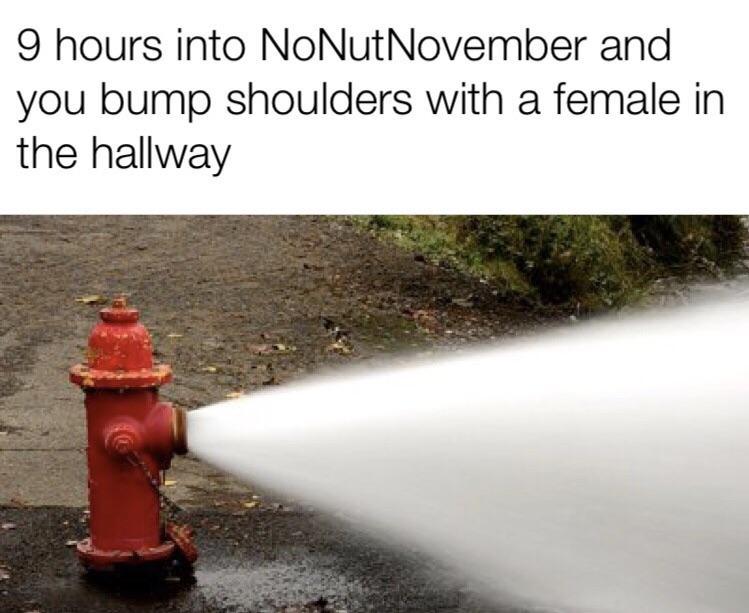}
\caption{Dark Humor: Yes\\Target: Gender/Sex\\Intensity: 3}
\label{fig:img3}
\end{subfigure}
\hfill
\begin{subfigure}{0.24\textwidth}
\includegraphics[width=\linewidth, height=3.5cm]{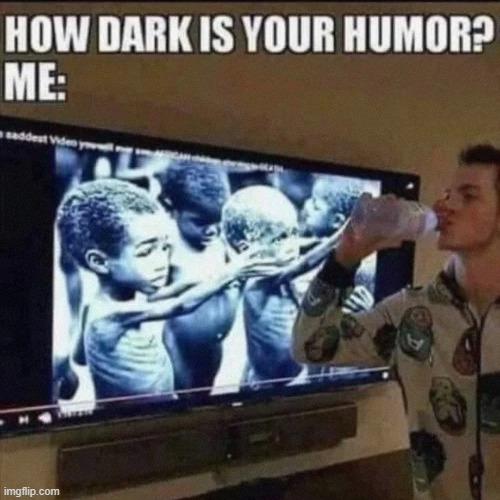}
\caption{Dark Humor: Yes\\Target: Race/Ethnicity\\Intensity: 3}
\label{fig:img4}
\end{subfigure}

\vspace{0.5em} 

\begin{subfigure}{0.24\textwidth}
\includegraphics[width=\linewidth, height=3.5cm]{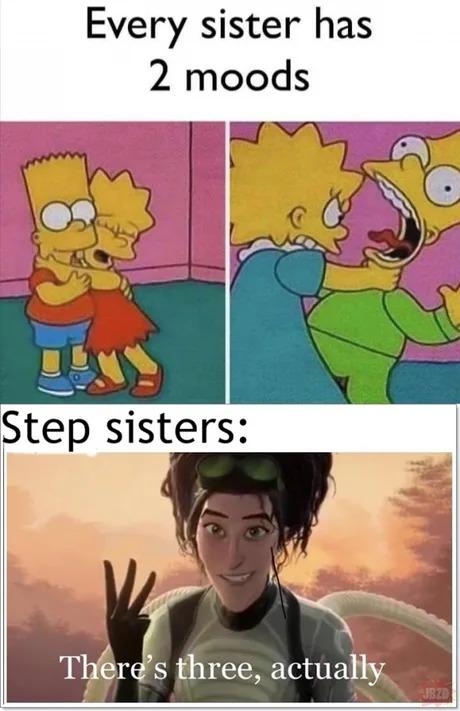}
\caption{Dark Humor: Yes\\Target: Gender/Sex\\Intensity: 2}
\label{fig:img5}
\end{subfigure}
\hfill
\begin{subfigure}{0.24\textwidth}
\includegraphics[width=\linewidth, height=3.5cm]{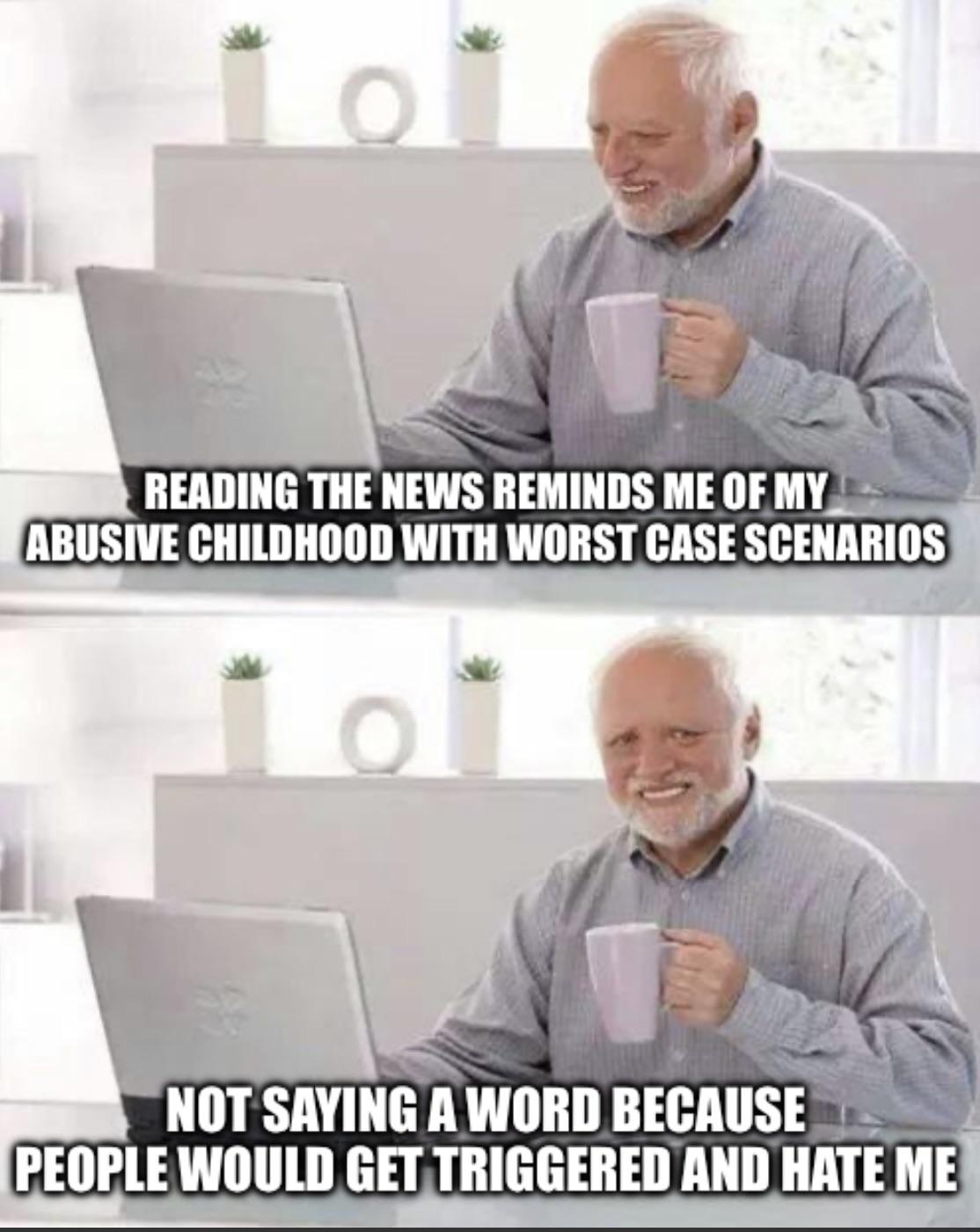}
\caption{Dark Humor: Yes\\Target: Mental Health\\Intensity: 2}
\label{fig:img6}
\end{subfigure}
\hfill
\begin{subfigure}{0.24\textwidth}
\includegraphics[width=\linewidth, height=3.5cm]{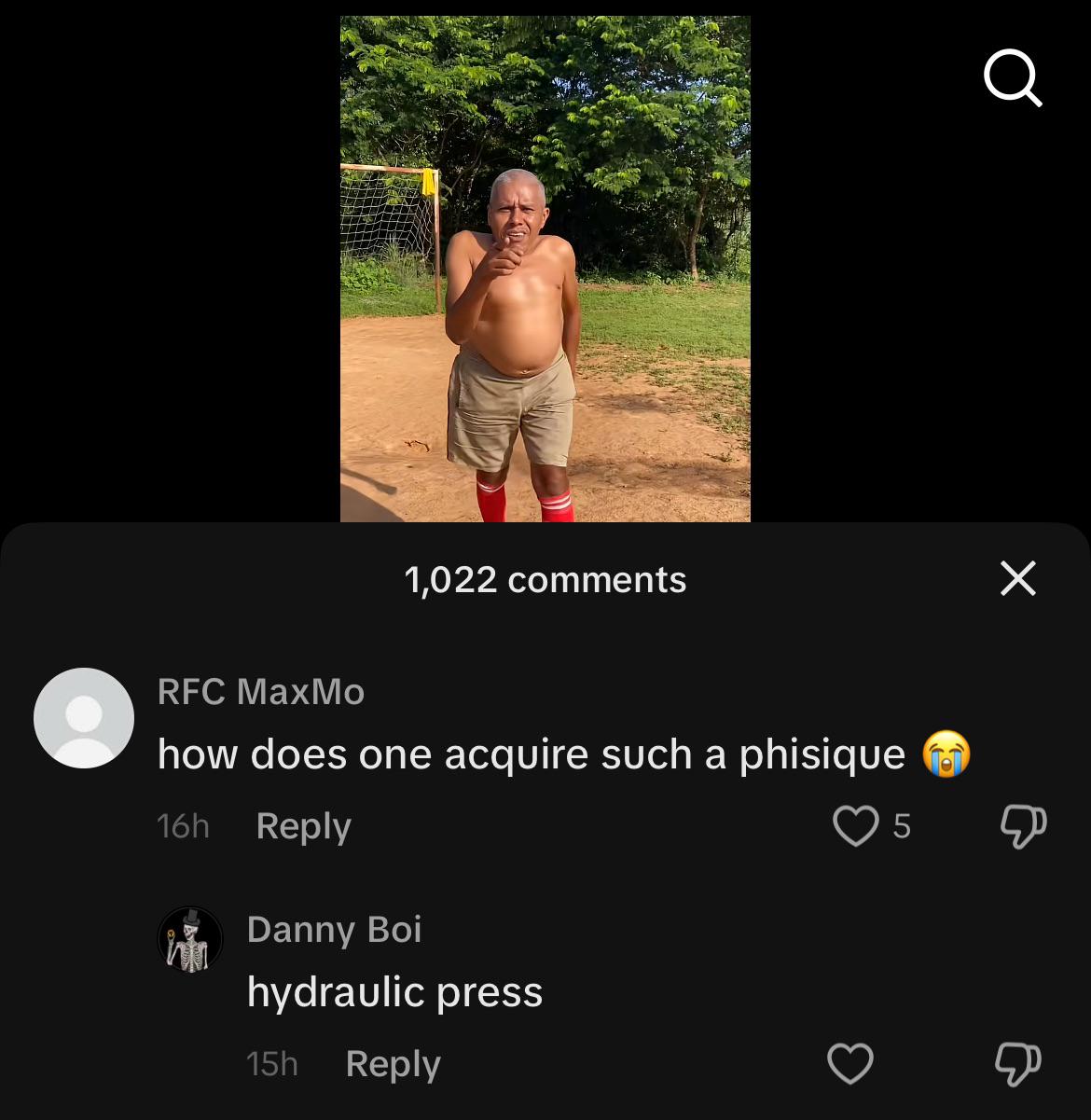}
\caption{Dark Humor: Yes\\Target: Disability\\Intensity: 2}
\label{fig:img7}
\end{subfigure}
\hfill
\begin{subfigure}{0.24\textwidth}
\includegraphics[width=\linewidth, height=3.5cm]{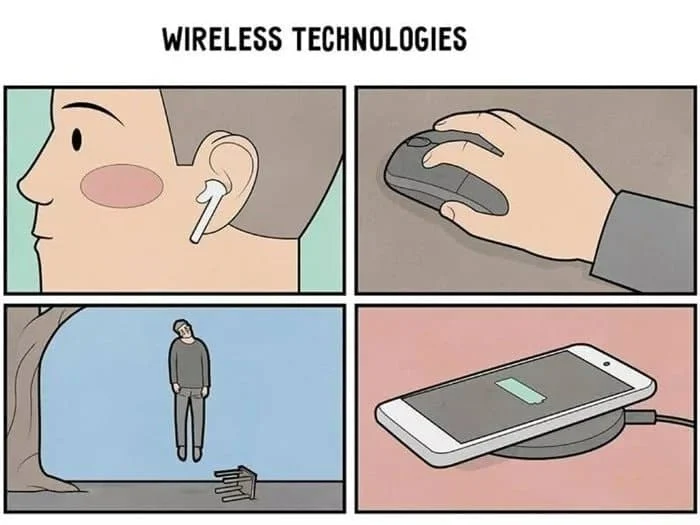}
\caption{Dark Humor: Yes\\Target: Violence/Death\\Intensity: 3}
\label{fig:img8}
\end{subfigure}

\caption{A subset of memes from D-Humor dataset, classified by their dark humor presence, target, and intensity. (Source: Reddit)}
\label{fig:darkhumor_grid}
\end{figure*}

\subsection{Dataset Collection}
All memes in our dataset were sourced from Reddit. We used the Reddit API to retrieve posts containing both an image and accompanying text, and applied OCR to extract embedded text. A subsequent manual screening eliminated the duplicates, resulting in a high-quality corpus of 4,379 multimodal memes for annotation and analysis.

\subsection{Annotator Training}
The annotation process was overseen by a faculty member and a PhD researcher with expertise in computational humor and harmful content analysis. The actual annotations were carried out by three undergraduate computer science students who voluntarily participated with informed consent. As a token of appreciation, they were granted access to A100 GPU resources for their own research.

A gold-standard seed set of 100 memes, balanced across dark and non-dark humor, was first annotated by the expert team. This dataset, along with a detailed codebook, was used to train the student annotators. The codebook defined annotation criteria with examples for three subtasks: (i) dark humor identification, (ii) target of the humor, and (iii) intensity level. After the initial labeling round, annotation discrepancies were discussed collectively to refine interpretation and improve consistency.

\subsubsection{Annotation Dimensions}
Each meme was annotated across three distinct dimensions:

\begin{itemize}
    \item \textbf{Dark Humor (DH)}: Binary label indicating whether the meme contains dark humor.
    \item \textbf{Target}: The specific group, topic, or subject that is the focus of the meme’s humor (e.g., gender, mental health, disability).
    \item \textbf{Intensity}: The perceived intensity of the dark humor, annotated on a three-level ordinal scale:
    \begin{itemize}
        \item \textit{Mild (1)}: Humor that is subtly dark or lightly suggestive.
        \item \textit{Moderate (2)}: Humor that is more direct and clearly involves a sensitive theme.
        \item \textit{Severe (3)}: Highly explicit or extreme dark humor involving graphic or taboo content.
    \end{itemize}
\end{itemize}

\subsection{Annotation in Batch Mode}
Following the training phase, the remaining memes were annotated in weekly batches of 200 memes. To mitigate potential cognitive fatigue from exposure to sensitive content, annotators were encouraged to take breaks of 10–15 minutes after each set of 10 memes, aligning with recommendations in prior work on annotator well-being~\cite{ybarra2006examining}. We also enforced a soft cap of 50 memes per day to prevent overload. Regular check-in meetings were held to discuss any difficult cases, resolve annotation conflicts, and monitor the psychological impact of labeling dark content~\cite{rehman2025implihatevid}. All annotations were finally reviewed by the supervising researchers to create a high-quality, adjudicated gold-standard dataset.

\subsection{Inter-Annotator Agreement}

To assess the consistency of annotations across annotators, we computed inter-annotator agreement scores using both pairwise Cohen’s Kappa and Fleiss’ Kappa. \autoref{tab:agreement_scores} summarizes the results for the three annotation tasks: \textit{Dark Humor} (DH), \textit{Target}, and \textit{Intensity}.

For the dark humor task, unweighted Cohen’s Kappa scores ranged from 58.92 to 85.88 across annotator pairs, with a Fleiss’ Kappa of 70.29, indicating substantial agreement. The target task yielded similarly strong results, with pairwise Cohen’s Kappa scores as high as 83.55 and a Fleiss’ Kappa of 72.13, reflecting high agreement on identifying the meme's intended subject.

\begin{table}[htbp]
\caption{Inter-annotator agreement scores. \textbf{A}: Annotator; \textbf{DH}: Dark Humor; \textbf{Unwt}: Unweighted; \textbf{Wt}: Weighted; \textbf{C Kappa}: Cohen's Kappa; \textbf{Wt C Kappa}: Weighted Cohen's Kappa; \textbf{F Kappa}: Fleiss' Kappa.}
\begin{center}
\begin{tabular}{|c|c|c|c|c|}
\hline
\textbf{Task} & \textbf{Ann Pair} & \textbf{UnWt C Kappa} & \textbf{Wt C Kappa} & \textbf{F Kappa} \\
\hline
\multirow{3}{*}{DH} 
& A1\&A2 & 66.05 & - & \multirow{3}{*}{70.29} \\
& A2\&A3 & 85.88 & - &  \\
& A3\&A1 & 58.92 & - &  \\
\hline
\multirow{3}{*}{Target} 
& A1\&A2 & 70.31 & - & \multirow{3}{*}{72.13} \\
& A2\&A3 & 83.55 & - &  \\
& A3\&A1 & 62.47 & - &  \\
\hline
\multirow{3}{*}{Intensity} 
& A1\&A2 & 53.24 & 74.39 & \multirow{3}{*}{58.57} \\
& A2\&A3 & 76.59 & 87.26 &  \\
& A3\&A1 & 45.89 & 69.45 &  \\
\hline
\end{tabular}
\label{tab:agreement_scores}
\end{center}
\end{table}

The intensity task, which involves ordinal labels (mild, moderate, severe), naturally presents more subjectivity. We computed both unweighted and \textbf{weighted} Cohen’s Kappa for this task only, since weighted Kappa accounts for the degree of disagreement in ordinal annotations. While unweighted scores ranged from 45.89 to 76.59, the weighted Kappa values were significantly higher, reaching up to 87.26. The Fleiss’ Kappa for intensity was 58.57, suggesting moderate agreement overall.

 The use of annotator training, expert supervision, and structured annotation guidelines helped ensure consistent and replicable labeling across complex and sensitive humor categories.

\begin{table*}[htbp]
\caption{Summary of dataset}
\begin{center}
\renewcommand{\arraystretch}{1.2}
\setlength{\tabcolsep}{8pt}
\begin{tabular}{|c|c||c|c|}
\hline
\multicolumn{2}{|c||}{\textbf{Train Data}} & \multicolumn{2}{c|}{\textbf{Test Data}} \\
\hline
\textbf{Dark Humor} & \textbf{Count} & \textbf{Dark Humor} & \textbf{Count} \\
\hline
No & 1,926 & No & 479 \\
Yes & 1,577 & Yes & 397 \\
\hline
\end{tabular}

\vspace{0.6cm}

\begin{tabular}{|l|cccc||l|cccc|}
\hline
\multicolumn{5}{|c||}{\textbf{Train Data: Target vs Intensity}} & \multicolumn{5}{c|}{\textbf{Test Data: Target vs Intensity}} \\
\hline
\textbf{Target} & \textbf{1} & \textbf{2} & \textbf{3} & \textbf{Total} & \textbf{Target} & \textbf{1} & \textbf{2} & \textbf{3} & \textbf{Total} \\
\hline
Gender/Sex-Related Topics & 303 & 248 & 132 & 683 & Gender/Sex-Related Topics & 59 & 74 & 28 & 161 \\
Mental Health & 169 & 49 & 4 & 222 & Mental Health & 46 & 9 & 3 & 58 \\
Disability & 30 & 26 & 19 & 75 & Disability & 9 & 7 & 3 & 19 \\
Race/Ethnicity & 32 & 35 & 19 & 86 & Race/Ethnicity & 5 & 7 & 6 & 18 \\
Violence/Death & 94 & 48 & 18 & 160 & Violence/Death & 31 & 13 & 3 & 47 \\
Other & 275 & 60 & 16 & 351 & Other & 79 & 12 & 3 & 94 \\
\hline
\textbf{Total} & \textbf{903} & \textbf{466} & \textbf{208} & \textbf{1,577} & \textbf{Total} & \textbf{229} & \textbf{132} & \textbf{46} & \textbf{397} \\
\hline
\end{tabular}
\end{center}
\label{tab:darknet_summary}
\end{table*}

\subsection{Dataset Statistics}

\autoref{tab:darknet_summary} presents a detailed breakdown of our dataset across both training and test splits. The dataset consists of meme samples labeled for the presence or absence of dark humor, along with fine-grained annotations on their target category and humor intensity. A sample of memes from the D-HUMOR dataset is shown in \autoref{fig:darkhumor_grid}.

\subsubsection{\textbf{Dark Humor Distribution}} 
The training set contains a total of 3,503 memes, with 1,577 labeled as containing dark humor and 1,926 as non-dark humor. The test set comprises 876 memes, including 397 positive and 479 negative samples. This relatively balanced distribution supports effective supervised training and facilitates fair evaluation.

\subsubsection{\textbf{Target Categories}}

Each meme labeled as dark humor was further annotated with the specific \textit{target} of its humor. In the training set, the most frequently targeted categories are \textit{Gender/Sex-Related Topics} (683 instances), followed by \textit{Other} (351), and \textit{Mental Health} (222). These three categories also dominate the test set distribution, indicating consistent prevalence across both data splits.

\subsubsection{\textbf{Intensity Levels}}

In the training data, the distribution is skewed toward lower intensity: 903 memes are rated as \textit{Mild}, 466 as \textit{Moderate}, and only 208 as \textit{Severe}. A similar trend appears in the test set, with 229, 132, and 46 memes across the three levels, respectively. This distribution reflects the real-world observation that while most dark humor is implicit or moderately suggestive, highly explicit content is comparatively rare.

Overall, this rich annotation framework provides valuable granularity for developing and benchmarking models that aim to understand not only the presence of dark humor but also its target and intensity.

\section{Methodology}
\begin{figure*}
    \centering
    \includegraphics[width=1\linewidth]{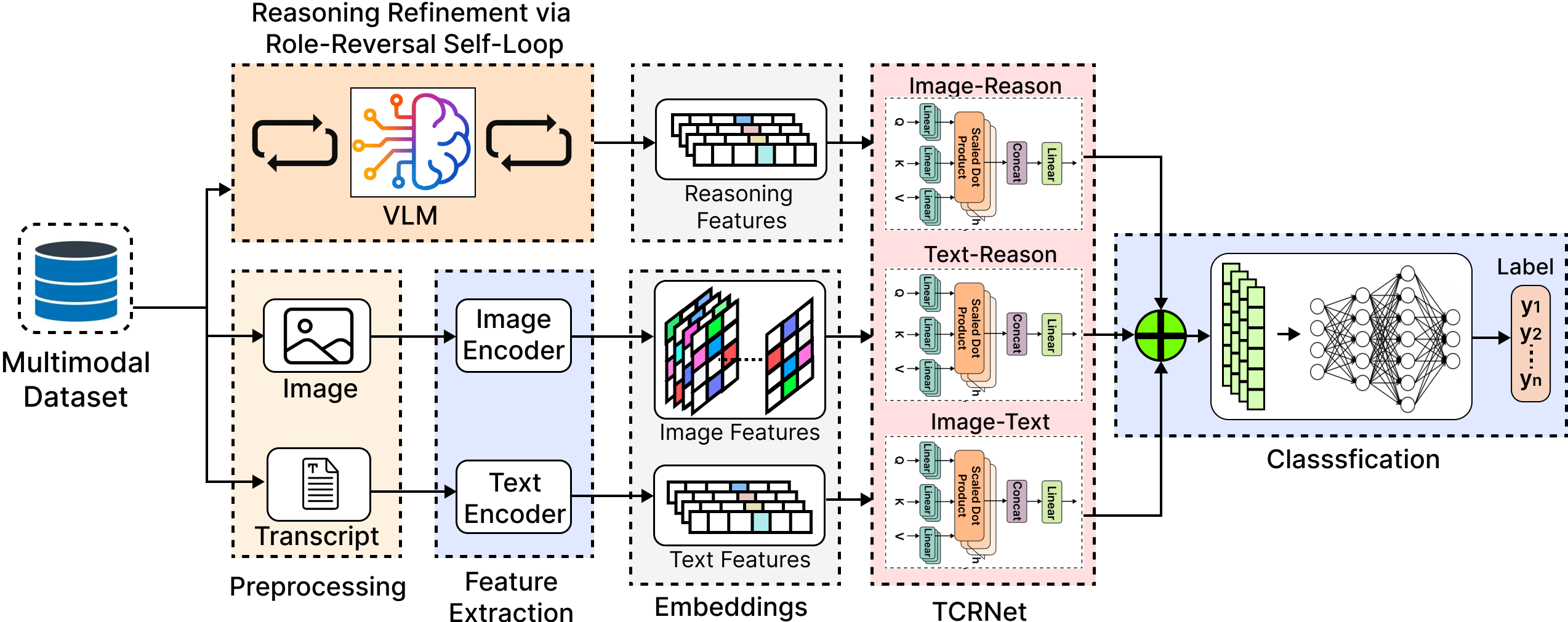}
    \caption{Overview of the proposed reasoning‐augmented multimodal architecture. Each meme (image + OCR text) is first processed by VLM to generate and refine a structured explanation via the Role-Reversal Self-Loop mechanism. Text encoders encode the OCR transcript and the self-refined explanation into textual embeddings, while the image encoder extracts visual embeddings from the meme image. A Tri‐stream Cross‐Reasoning Network (TCRNet) then applies pairwise scaled dot-product attention across the three modalities, text, image, and reasoning, to produce attended representations. These representations are concatenated into a unified feature vector, which is passed through a classification head to predict dark humor presence, target category, and intensity.}
    \label{fig:method}
\end{figure*}

Understanding dark humor in memes requires reasoning beyond literal content, as it often relies on abstract incongruity between text and visuals. To address this, we propose a reasoning-augmented multimodal framework that leverages linguistic cues, image semantics, and inferred commonsense reasoning. The architecture, as shown in \autoref{fig:method}, comprises five components: reasoning refinement, feature extraction, cross-attention fusion, feature aggregation, and final classification.

\subsection{Structured Reasoning Generation}

Understanding dark humor memes requires more than just identifying surface-level content. The humor often hinges on subtle context, cultural subtext, visual incongruities, and emotionally conflicting cues, many of which may be implicit, taboo, or deliberately ambiguous. These characteristics make dark humor memes uniquely challenging to interpret, both for humans and machines.

To address this, we enrich the meme classification process with structured explanatory signals. By explicitly breaking down the reasoning behind a meme's humor, emotional tone, and target, we provide the model with additional context that aids in disambiguation. This additional supervision helps the model move beyond superficial patterns and capture the deeper intent and affective contrast that are typical in dark humor. Moreover, structured reasoning allows for improved transparency and interpretability, supporting downstream applications like content moderation, harm detection, and bias analysis.

\subsubsection{\textbf{Structured Explanation Schema}}
We design each explanation using a six-part schema that captures distinct dimensions of the meme's communicative intent and stylistic choices:
\begin{enumerate}
    \item \textbf{Meme Summary:} A concise description of the meme’s image and text.
    \item \textbf{Implied Joke:} The core humorous element inferred from the meme.
    \item \textbf{Narrative Structure:} The rhetorical style (e.g., sarcasm, irony, absurdism).
    \item \textbf{Emotional Effect:} The emotional response the meme is designed to elicit.
    \item \textbf{Dark Attributes:} Any presence of taboo, offensive, or socially sensitive themes.
    \item \textbf{Target:} The entity, group, or concept being referenced or mocked.
\end{enumerate}

This decomposition facilitates fine-grained understanding, and the modular format can be encoded and fused with vision-language features during classification.

\subsubsection{\textbf{Iterative Reasoning Refinement via Role-Reversal Self-Loop}}

Interpreting dark humor requires not only visual-textual understanding but also alignment with nuanced social, cultural, and emotional cues, an area where even large models often fail. To bridge this gap, we propose a novel technique called \textbf{Role-Reversal Self-Loop}, a simple yet powerful mechanism for enhancing \textit{AI alignment} in tasks involving implicit reasoning and culturally sensitive content.

\begin{figure}[htbp]
    \centering
    \includegraphics[width=0.99\linewidth]{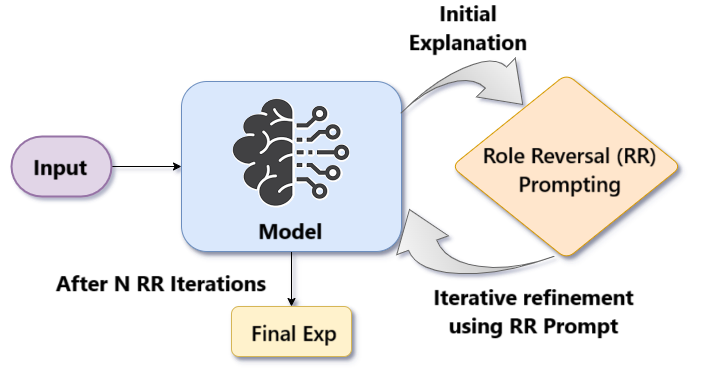}
    \caption{
        \textbf{Self-loop mechanism using Role Reversal (RR) Prompting.}
        The input meme is passed to a language model which generates an initial explanation. This explanation is then refined iteratively through Role Reversal prompting, where the model assumes the role of a reviewer to critique and improve its own output. After \( N \) such refinement cycles, a final explanation is obtained. This mechanism enables the model to simulate self-awareness and improve explanatory coherence without external supervision.
    }
    \label{fig:self-loop-rr}
\end{figure}

After generating an initial structured reasoning using the Qwen-2.5-7B Vision-Language Model (VLM), we prompt the model to assume a reversed role: \textit{it becomes the original author of the meme}. Under this setup, the model is asked to review the reasoning it “received” for its own meme and critically assess whether the fields are faithful and complete as shown in \autoref{fig:self-loop-rr}. This role-flip is psychologically aligned with reflective human behavior, encouraging the model to evaluate explanations with a self-interested perspective, thereby improving alignment.

This approach has two key benefits:
\begin{itemize}
    \item It ensures that each explanation component is reviewed in context, preventing the model from over-focusing on only one field (e.g., the joke or the target).
    \item It encourages introspective correction without requiring external feedback or manual interventions, thereby improving scalability and automation.
\end{itemize}

We apply this refinement loop iteratively for a maximum of 5 epochs. On average, explanations converge and stabilize by approximately 3 epochs, indicating that the model typically needs only a few rounds to self-correct and align its output. Importantly, our manual review of a subset of explanations revealed that most improvements were concentrated in the fields of Narrative Structure, Dark Attributes, and Implied Joke, components that demand deeper contextual reasoning. Fields like Meme Summary, Target, and Emotional Effect were more stable across iterations, suggesting they are more straightforward to interpret initially.

\begin{tcolorbox}[colback=gray!10, colframe=black!80, title=Role-Reversal Self-Loop Prompt Template]

\hspace{2.5mm}You are the author and creator of the given dark humor meme.
Someone else has provided a detailed explanation of your meme’s meaning, humor, and components, including the Meme Summary, Implied Joke, Narrative Structure, Emotional Effect, Dark Attributes, and Target.\\

\hspace{2.5mm}Your task is to review their explanation from your perspective as the original author. Analyze how well their reasoning aligns with your intended humor, message, tone, and overall context. For each component, identify any inaccuracies, missing details, or misunderstandings. Then, provide a thorough, corrected explanation that fully reflects your original intent, ensuring clarity, accuracy, and completeness.

Please present your revised explanation below.\\

\small \texttt{[Explanation Components Inserted Here]}
\end{tcolorbox}

\vspace{1mm}

\subsection{Feature Extraction}

We extract embeddings from three modalities: the meme’s original transcript $T$, the generated reasoning $R$, and the image $I$.

\vspace{2mm}
\noindent\textbf{Textual Features:}
We use the BERT model to encode both $T$ and $R$ into contextual token embeddings:
\begin{align}
    \mathbf{F}_T &= \text{BERT}(T) \in \mathbb{R}^{197 \times 768} \\
    \mathbf{F}_R &= \text{S-BERT}(R) \in \mathbb{R}^{197 \times 768}
\end{align}
Each row corresponds to a subword token in a 768-dimensional embedding space.

\vspace{2mm}
\noindent\textbf{Visual Features:}
We use the ViT model, which combines vision and language pretraining objectives, to extract features from the meme image:
\begin{equation}
    \mathbf{F}_I = \text{ViT}(I) \in \mathbb{R}^{197 \times 768}
\end{equation}
This representation captures visual cues such as objects, expressions, and contextual artifacts relevant to the meme.

\subsection{Tri-stream Cross-Reasoning Network (TCRNet)}

To effectively model inter-modal interactions and capture reasoning-driven dependencies, we introduce a \textit{Tri-stream Cross-Reasoning Network} \cite{zhou2021multimodal, tang2022triple}. This module enables each modality, text (\(T\)), image (\(I\)), and LLM-generated reasoning (\(R\)), to mutually inform and align with one another by attending across pairs.

Specifically, we use scaled dot-product attention to implement pairwise reasoning flows. Let $\mathbf{Q}_X = \mathbf{F}_X \mathbf{W}_Q$, $\mathbf{K}_Y = \mathbf{F}_Y \mathbf{W}_K$, and $\mathbf{V}_Y = \mathbf{F}_Y \mathbf{W}_V$ be the linear projections of modalities $X$ and $Y$, where \(X, Y \in \{T, I, R\}\). The attention from $X$ to $Y$ is computed as:
\begin{equation}
    \text{Attn}(X \rightarrow Y) = \text{softmax} \left( \frac{\mathbf{Q}_X \mathbf{K}_Y^\top}{\sqrt{d_k}} \right) \mathbf{V}_Y
\end{equation}

This formulation allows modality \(X\) to selectively extract context-aware signals from modality \(Y\), thereby modeling semantic incongruity, visual-verbal contradictions, and inferential alignment, all of which are critical in dark humor.

We compute such cross-reasoning across all three modality pairs:
\begin{align}
    \widetilde{\mathbf{F}}_{T \leftrightarrow I} &= \text{Attn}(T \rightarrow I) \\
    \widetilde{\mathbf{F}}_{T \leftrightarrow R} &= \text{Attn}(T \rightarrow R) \\
    \widetilde{\mathbf{F}}_{I \leftrightarrow R} &= \text{Attn}(I \rightarrow R)
\end{align}

\subsection{Feature Aggregation}

Each cross-reasoned output is aggregated using average pooling to yield fixed-length vectors:
\begin{align}
    \mathbf{v}_{T \leftrightarrow I} &= \text{AvgPool}(\widetilde{\mathbf{F}}_{T \leftrightarrow I}) \\
    \mathbf{v}_{T \leftrightarrow R} &= \text{AvgPool}(\widetilde{\mathbf{F}}_{T \leftrightarrow R}) \\
    \mathbf{v}_{I \leftrightarrow R} &= \text{AvgPool}(\widetilde{\mathbf{F}}_{I \leftrightarrow R})
\end{align}

We concatenate the resulting vectors to obtain a unified representation:
\begin{equation}
    \mathbf{v} = [\mathbf{v}_{T \leftrightarrow I}; \mathbf{v}_{T \leftrightarrow R}; \mathbf{v}_{I \leftrightarrow R}] \in \mathbb{R}^{2,304}
\end{equation}
This vector serves as the multimodal reasoning-aware representation fed into the final classification layer.

\subsection{Hyperparameters and System Description}
\autoref{tab:hyperparams} lists the primary hyperparameters used for the proposed framework. Encoder-only text and image models share identical training configurations, while LLMs fine-tuned with QLoRA use 3 epochs and low-rank adaptation settings ($r=8$, $\alpha=32$, dropout 0.1). The TCRNet additionally employs dropout (0.3) and eight attention heads in its training setup.

\begin{table}[ht]
\centering
\caption{Key hyperparameter configurations used across different model types.}
\label{tab:hyperparams}
\begin{tabular}{|l|l|l|}
\hline
\textbf{Model Type} & \textbf{Hyperparameter} & \textbf{Value} \\
\hline
\multirow{3}{*}{Encoder-Only Text Models} 
    & Epochs & 5 \\
    & Batch Size & 16 \\
    & Learning Rate & 0.00002 \\
\hline
\multirow{4}{*}{LLMs with QLoRA} 
    & Epoch & 3 \\
    & LoRA Rank ($r$) & 8 \\
    & LoRA Alpha & 32 \\
    & LoRA Dropout & 0.1 \\
    & Target Modules & q\_proj, v\_proj \\
\hline
\multirow{3}{*}{Image Models} 
    & Epochs & 5 \\
    & Batch Size & 16 \\
    & Learning Rate & 0.00002 \\
\hline
\multirow{6}{*}{TCRNet} 
    & Epochs & 5 \\
    & Batch Size & 16 \\
    & Learning Rate & 0.00002 \\
    & Dropout & 0.3 \\
    & Attention Heads & 8 \\
\hline
\end{tabular}
\end{table}

All experiments were conducted on a server equipped with Intel(R) Xeon(R) Silver 4316 CPU @ 2.30GHz, 4× NVIDIA A16 GPUs (each with 16 GB VRAM), running CUDA 12.8 and driver version 570.133.07. The training utilized PyTorch with GPU acceleration, and computations were distributed using a single GPU unless specified otherwise.

\section{Results and Discussion}

\begin{table*}[htbp]
\caption{Evaluation of various models across different input modalities for dark humor meme analysis. The highest performance in each metric is highlighted in \textbf{bold}, while the second-best performance is marked with a \textbf{$^\dagger$} symbol. \textbf{P-corr} denotes the Pearson correlation coefficient used to evaluate the ordinal classification task for Intensity.}
\begin{center}
\begin{tabular}{|l|ccc|ccc|cccc|}
\hline
\textbf{Model} 
& \multicolumn{3}{c|}{\textbf{Dark Humor (Yes/No)}} 
& \multicolumn{3}{c|}{\textbf{Target}} 
& \multicolumn{4}{c|}{\textbf{Intensity}} \\
\cline{2-11}
& \textbf{Accuracy} & \textbf{Macro-F1} & \textbf{Wt-F1} 
& \textbf{Accuracy} & \textbf{Macro-F1} & \textbf{Wt-F1} 
& \textbf{Accuracy} & \textbf{Macro-F1} & \textbf{Wt-F1} & \textbf{P-Corr.} \\
\hline
\multicolumn{11}{|l|}{\textbf{OCR Text Only}} \\
\hline
BERT\cite{devlin2019bert} & 70.05 & 69.41 & 69.75 & 59.69 & 53.27 & 59.53 & 58.91$^\dagger$ & 46.59$^\dagger$ & 57.63$^\dagger$ & 27.18 \\
Distil-BERT\cite{sanh2019distilbert} & 69.81 & 69.23 & 69.55 & 61.50 & 55.98 & 60.91 & 55.81 & 45.75 & 55.22 & 26.25 \\
Llama-3.2-3b-it\cite{touvron2023llama} & 71.72 & 71.14 & 71.45 & 54.52 & 38.20 & 51.96 & 57.62 & 37.41 & 52.58 & 17.05 \\
\hline
\multicolumn{11}{|l|}{\textbf{Image Only}} \\
\hline
CLIP\cite{radford2021learning} & 57.76 & 57.69 & 57.85 & 37.53 & 35.43 & 38.47 & 40.30 & 31.18 & 40.72 & 2.40 \\
ViT\cite{dosovitskiyimage} & 64.04 & 62.82 & 63.45 & 45.59 & 32.76 & 39.86 & 54.66 & 36.22 & 51.14 & 7.14 \\
BLIP\cite{li2022blip}  & 64.95 & 64.25 & 64.72 & 44.84 & 40.53 & 43.79 & 58.19 & 41.69 & 54.40 & 19.96 \\
\hline
\multicolumn{11}{|l|}{\textbf{VLMs (Prompt based Zero-Shot)}} \\
\hline
Qwen-2.5-7b-it\cite{b18} & 64.27 & 63.26 & 63.83 & 57.68 & 53.70 & 56.28 & 17.63 & 15.27 & 14.14 & 17.50 \\
LLaVA-1.6-mistral-7b\cite{liu2023visual} & 45.32 & 31.19 & 28.27 & 48.87 & 48.30 & 49.14 & 30.23 & 24.22 & 29.92 & 11.48 \\
\hline
\multicolumn{11}{|l|}{\textbf{OCR Text + Structured Explanation (with Role-Reversal Prompting)}} \\
\hline
BERT & 71.24 & 70.88 & 71.13 & 65.37$^\dagger$ & 61.67$^\dagger$ & 65.42$^\dagger$ & 56.33 & 43.16 & 55.10 & 32.12 \\
Distil-BERT & 70.53 & 70.02 & 70.32 & \textbf{66.41} & \textbf{62.53} & \textbf{66.07} & 55.56 & 43.98 & 54.95 & 32.56$^\dagger$ \\
Llama-3.2-3b-it & 73.39$^\dagger$ & 72.91$^\dagger$ & 73.19$^\dagger$ & 59.17 & 50.37 & 57.62 & 58.14 & 37.94 & 53.76 & 24.4 \\
\hline
\multicolumn{11}{|l|}{\textbf{VLMs (Prompt based Zero-Shot with Role-Reversal Prompting)}} \\
\hline
Qwen-2.5-7b-it & 64.04 & 64.02 & 63.95 & 55.67 & 50.97 & 55.01 & 29.22 & 28.97 & 29.33 & 11.84 \\
LLaVA-1.6-mistral-7b & 46.80 & 35.40 & 32.86 & 49.62 & 51.44 & 50.03 & 37.03 & 34.41 & 39.35 & 15.22 \\
\hline

\textbf{TCRNet (ours)} & \textbf{75.00} & \textbf{73.55} & \textbf{74.13} & 64.48 & 60.54 & 64.17 & \textbf{62.72} & \textbf{49.71} & \textbf{60.74} & \textbf{38.63} \\
\hline
\end{tabular}
\label{tab:results_comparison}
\end{center}
\end{table*}

\subsection{Experimental Findings}

\autoref{tab:results_comparison} presents a comprehensive evaluation of various models on three tasks: dark humor detection (binary classification), target identification (multi-class classification), and intensity prediction. We report Accuracy, Macro-F1, and Weighted-F1 for all tasks, and additionally include Pearson correlation for intensity prediction, as it is ordinal. When relying solely on OCR-extracted text, pretrained models like BERT, DistilBERT, and LLaMA achieve reasonable performance for dark humor detection. LLaMA leads with 71.72\% accuracy and a Macro-F1 of 71.14\%. However, these text-only systems struggle to identify the specific target of the humor and to gauge its intensity. For instance, DistilBERT’s target identification Macro-F1 is only 55.98\%, and BERT’s intensity Pearson correlation is a modest 27.18\%. These results indicate that while textual cues carry considerable information for detecting the presence of dark humor, they lack the broader context and implicit signals needed to pinpoint whom or what the joke is aimed at, and how severe it is.

Purely image-based models, such as CLIP, ViT, and BLIP,  underperform compared to text-only baselines. BLIP achieves the highest image-only dark humor accuracy at 64.95\%, but target identification accuracy remains below 45\% and intensity accuracy under 60\%. The Pearson correlation for intensity is especially low, for instance, 19.96\% for BLIP. This suggests that, although visual features capture facial expressions, color palettes, and scene context, they fail to convey the nuanced semantics that define dark humor. In other words, imagery alone lacks the explicit linguistic content that often carries the punchline or the taboo reference.

Zero-shot vision-language models, such as Qwen-VL and Next-LLaVA-VL, demonstrate limited effectiveness without task-specific fine-tuning. Qwen-VL achieves 64.27\% dark humor accuracy and a 17.50\% intensity correlation, while Next-LLaVA-VL lags further behind (45.32\% accuracy for humor detection, 11.48\% intensity correlation). Even when provided with structured explanations as prompts, these VLMs show only marginal improvements. For example, Next-LLaVA-VL’s intensity correlation rises from 11.48\% to 15.22\% with explanations, but remains far below the performance of models trained on text-plus-explanation inputs. These observations highlight that, although large VLMs contain rich pretraining knowledge, they still require supervised adaptation to handle the implicit, sensitive nature of dark humor.

Augmenting OCR text with structured explanations yields consistent and substantive gains across all tasks. Under the “OCR Text + Structured Explanation” setting, BERT’s dark humor accuracy increases from 70.05\% to 71.24\%, and DistilBERT’s target Macro-F1 jumps from 55.98\% to 62.53\%. Most notably, intensity Pearson correlation for DistilBERT improves from 26.25\% to 32.56\%. These results underline that structured explanations, capturing implied jokes, narrative structure, and emotional effect, supplement raw text with crucial contextual cues. Even LLaMA, which already performs well with text alone, sees its dark humor accuracy rise to 73.39\% when explanations are included, confirming that reasoning signals complement large language model outputs.

Our proposed \textbf{TCRNet} model, which implements a Tri-stream Cross-Reasoning Network to fuse OCR text, image features, and structured explanations, achieves the best overall results. TCRNet attains 75.00\% accuracy on dark humor detection and 62.72\% accuracy on intensity prediction, both the highest scores among all evaluated methods. By explicitly aligning visual and textual streams with reasoning-derived explanations, TCRNet captures implicit associations and semantic incongruities that single-modality or non-reasoning models miss. These improvements demonstrate that reasoning-aware, multimodal fusion is essential for accurately understanding and categorizing dark humor memes.

\begin{figure*}[!htbp]
\centering
\begin{subfigure}[t]{0.32\textwidth}
    \includegraphics[width=\linewidth]{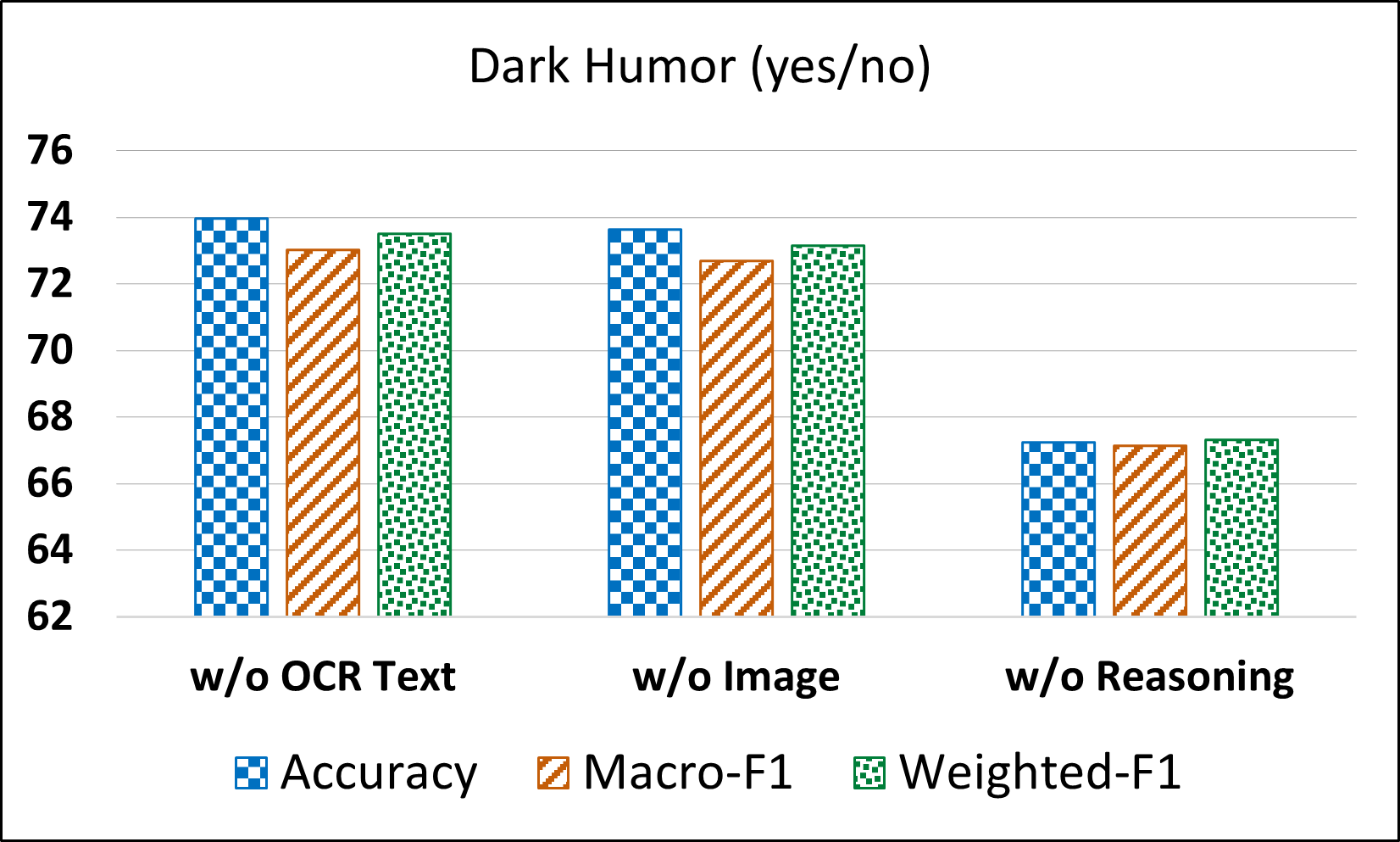}
    \caption{Dark humor}
    \label{fig:diag1}
\end{subfigure}
\hfill
\begin{subfigure}[t]{0.32\textwidth}
    \includegraphics[width=\linewidth]{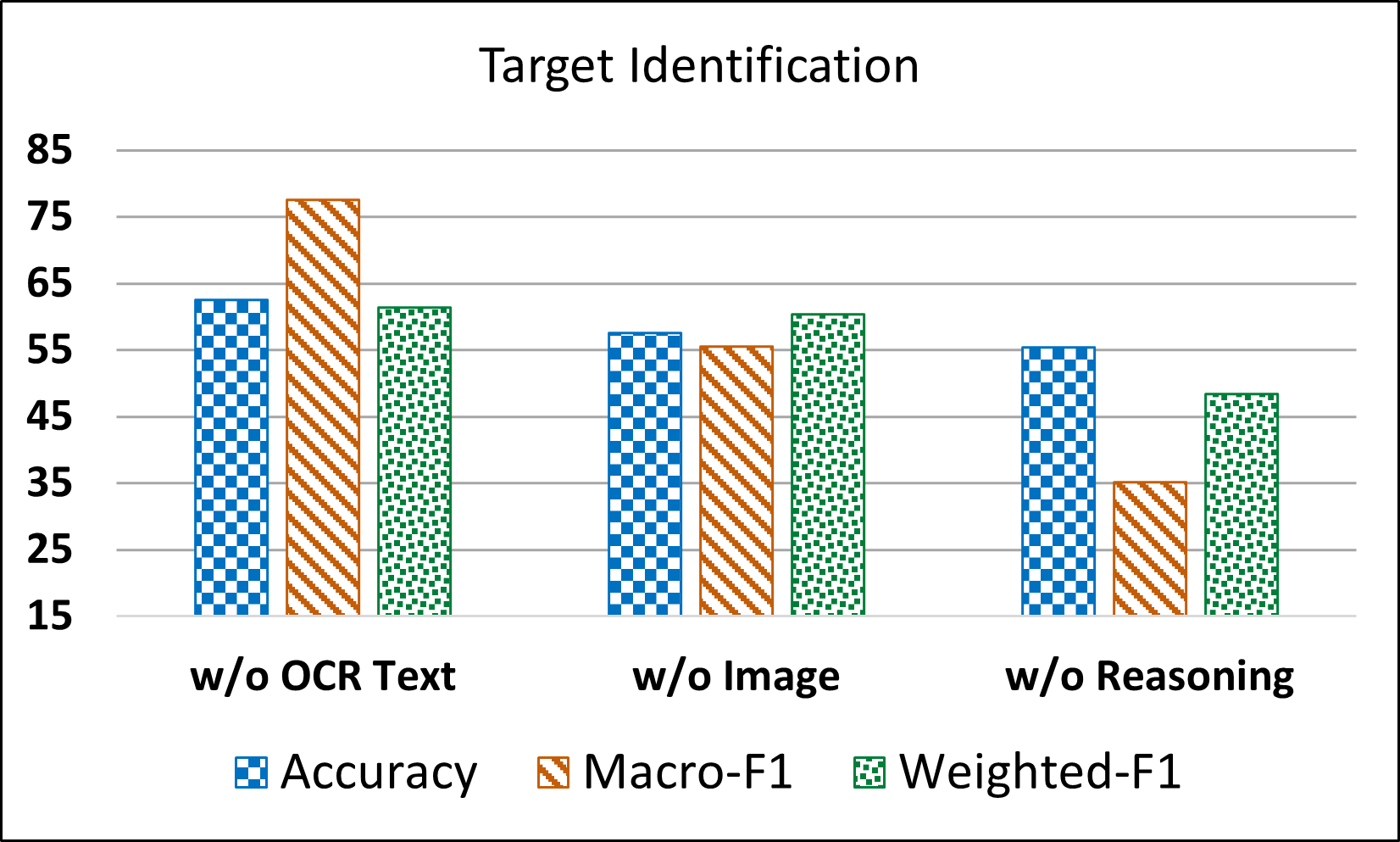}
    \caption{Target identification}
    \label{fig:diag2}
\end{subfigure}
\hfill
\begin{subfigure}[t]{0.32\textwidth}
    \includegraphics[width=\linewidth]{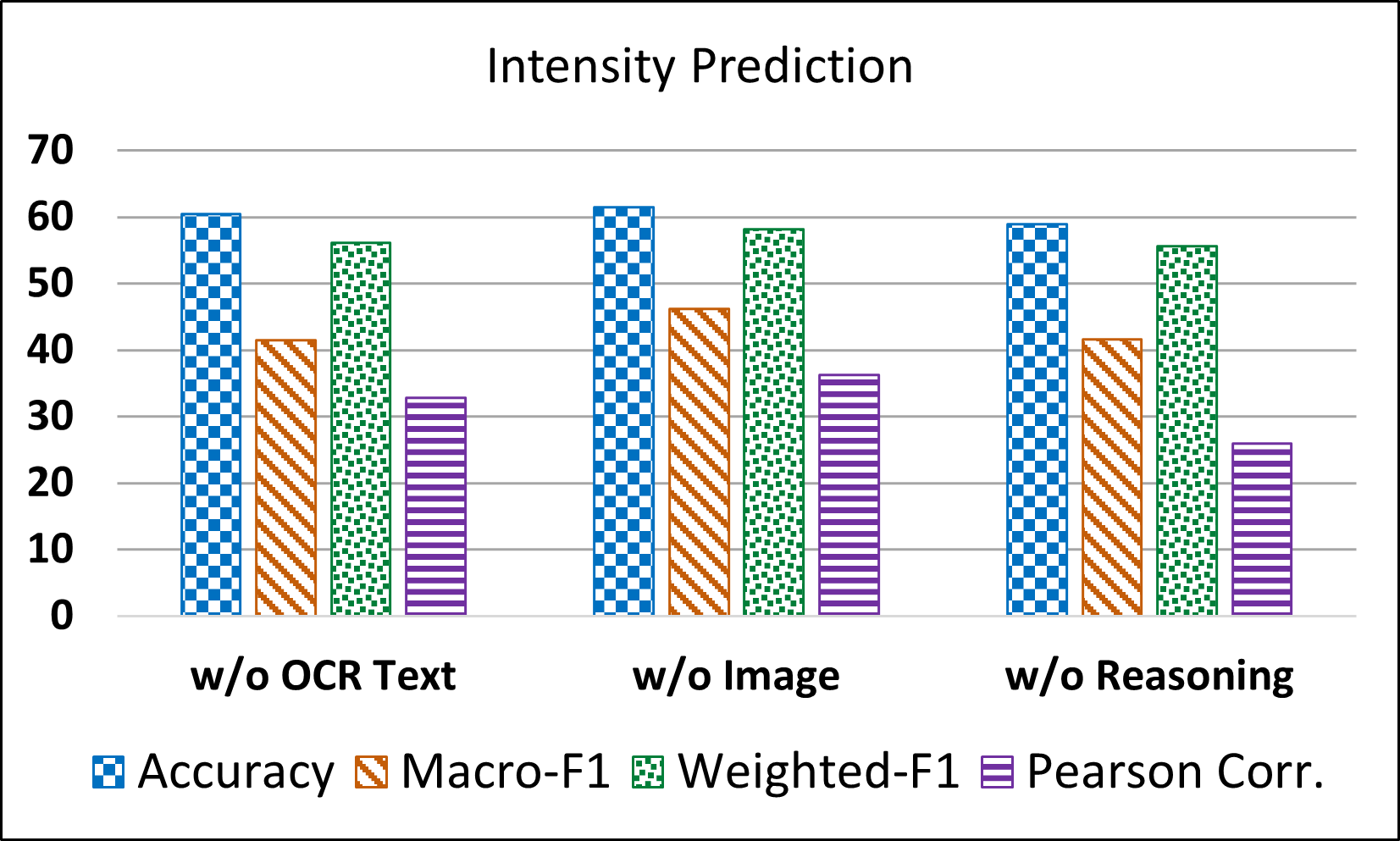}
    \caption{Intensity prediction}
    \label{fig:diag3}
\end{subfigure}
\caption{The charts represent an ablation study of TCRNet on Dark Humor Meme Analysis. Each ablation removes one component (OCR Text, Image, or Explanation) to study its impact. The full TCRNet results are included in \autoref{tab:results_comparison}.}
\label{fig:ablation}
\end{figure*}

\section{Ablation Study}
From the ablation study presented in \autoref{fig:ablation}, it is evident that the reasoning component based on structured explanations plays a critical role in enhancing the performance of TCRNet across all three tasks. When the explanation modality is removed, there is a substantial drop in performance, particularly in Target Identification (Macro-F1 drops from 60.54 to 35.11) and Dark Humor Classification (Weighted-F1 drops from 74.13 to 67.31). This clearly indicates that dark humor memes often rely on implicit, nuanced reasoning that goes beyond surface-level textual or visual cues.

Unlike typical classification problems, where raw OCR text or image features might suffice, dark humor detection demands an externalized and structured interpretive view. The explanation-based features help the model contextualize morally ambiguous content, sarcasm, or deception, which are often difficult to infer directly from the raw input. These findings emphasize the importance of integrating explicit reasoning mechanisms into multimodal understanding systems, especially for complex, subjective tasks like dark humor analysis.

\section{Conclusion}

In this work, we introduced the D-Humor dataset, a multimodal collection of 4,397 Reddit memes annotated for dark humor (yes/no), target category (gender, mental health, violence, race, disability, other), and a three-level intensity rating (mild, moderate, severe). We further proposed a reasoning-augmented framework that first generates and self-refines structured explanations using Qwen-2.5-7B VLM, then fuses OCR text, ViT-based image features, and refined reasoning via a Tri-stream Cross-Reasoning Network (TCRNet). Experimental results across three tasks, dark humor detection, target identification, and intensity prediction, demonstrate that (1) text-only or image-only models are insufficient for capturing the nuances of dark humor, (2) zero-shot vision-language models underperform without fine-tuning, and (3) augmenting OCR text with structured explanations yields consistent gains.  

Our proposed TCRNet achieves state-of-the-art performance, attaining 75.00\% accuracy on dark humor detection and 62.72\% accuracy on intensity prediction, outperforming all baselines by a significant margin. An ablation study confirms that the structured explanation component is critical: removing explanations leads to a dramatic drop in Target Identification Macro-F1 (from 60.54\% to 35.11\%) and Dark Humor Weighted-F1 (from 74.13\% to 67.31\%). These findings highlight that dark humor in memes relies on implicit, context-dependent cues, such as implied jokes and emotional effect, that cannot be inferred from raw text or images alone, necessitating explicit reasoning signals.

While we acknowledge that D-Humor is skewed toward the \textit{Gender/Sex-Related Topics} label, likely reflecting the greater availability of such content online, models trained on this data generalize well across all categories and do not exhibit overfitting. For future work, we plan to expand the dataset with additional target classes to enable finer-grained analyses, as well as investigate semi-supervised and self-supervised techniques to further improve generalization to emerging dark humor trends. We also aim to explore more advanced alignment mechanisms within the Tri-stream Cross-Reasoning framework to better capture evolving cultural and social nuances in online humor.

\section*{Ethical Consideration}
We propose the D-Humor Dataset as a novel resource to study dark humor in memes. Due to the sensitive and potentially offensive nature of the content, users should exercise caution and follow ethical guidelines when handling the data. The dataset is intended exclusively for academic and research purposes aimed at advancing understanding in this domain. To protect the privacy of individuals and maintain data integrity, public redistribution, sharing, or any unauthorized dissemination of the dataset is strictly prohibited. Access will be granted only upon signing a data usage agreement.

\section*{Reddit Ethical policy for scraping the memes}
In conducting our D-Humor Project, we collected publicly available memes from Reddit, ensuring strict adherence to Reddit's Privacy Policy and Public Content Policy.\footnote{\url{https://www.redditinc.com/policies/privacy-policy}} Reddit emphasizes user anonymity by not requiring real names or photos, stripping metadata from shared media, and prohibiting doxxing. Our data collection excluded any personally identifiable information, such as usernames, email addresses, or IP data, focusing solely on content that users have made publicly accessible. Reddit's Public Content Policy permits the use of public posts and comments for research, provided that user privacy is respected and content is not used for surveillance or identification.\footnote{\url{https://support.reddithelp.com/hc/en-us/articles/26410290525844}} We also ensured that no deleted or private content was included, aligning with Reddit's guidelines that prohibit unauthorized scraping and highlight the importance of user consent and data protection. We took care to follow the platform’s stated policies and norms regarding data usage.\footnote{\url{https://www.redditinc.com/policies/user-agreement-april-18-2023}}

\section*{Acknowledgment}
The authors thank IIT Indore for providing financial support for the registration of the article. The authors thank the UGC Junior Research Fellowship (UGC-JRF) scheme, an initiative of the Government of India, for providing the Ph.D. fellowship to Mohammad Zia Ur Rehman (NTA Ref. No.: 190510846524). The authors are also thankful for the Young Faculty Research Catalysing Grant (YFRCG) by the Indian Institute of Technology Indore for providing the resources to the project (Project ID: IITI/YFRCG/2023-24/03).

\balance
\bibliographystyle{IEEEtran}
\bibliography{references}

\end{document}